\documentclass{article}

    \usepackage[final]{neurips_2020}

\usepackage[utf8]{inputenc} 
\usepackage[T1]{fontenc}    
\usepackage{hyperref}       
\usepackage{url}            
\usepackage{booktabs}       
\usepackage{amsfonts}       
\usepackage{nicefrac}       
\usepackage{microtype}      
\usepackage{booktabs}
\usepackage{multirow}
\usepackage{graphicx}
\usepackage[table,xcdraw]{xcolor}

\title{CytoImageNet: A large-scale pretraining dataset for bioimage transfer learning}

\author{%
  Stanley Bryan Z. Hua \\
  Computer Science\\
  University of Toronto\\
  \texttt{\href{mailto:stanley.hua@mail.utoronto.ca}{stanley.hua@mail.utoronto.ca}}  
  \\
   \AND
   Alex X. Lu \\
   Computer Science\\
   University of Toronto\\
   (Current Affiliation: Microsoft Research)\\
   \texttt{lualex@microsoft.com} \\
   \And
   Alan M. Moses \\
   Cell and Systems Biology \\
   Computer Science \\
   Center of Genome Evolution and Function \\
   University of Toronto \\
   \texttt{alan.moses@utoronto.ca} \\
}

\begin{document}

\maketitle

\begin{abstract}
    Motivation: In recent years, image-based biological assays have steadily become high-throughput, sparking a need for fast automated methods to extract biologically-meaningful information from hundreds of thousands of images. Taking inspiration from the success of ImageNet, we curate CytoImageNet, a large-scale dataset of openly-sourced and weakly-labeled microscopy images (890K images, 894 classes). Pretraining on CytoImageNet yields features that are competitive to ImageNet features on downstream microscopy classification tasks. We show evidence that CytoImageNet features capture information not available in ImageNet-trained features. The dataset is made available at \url{https://www.kaggle.com/stanleyhua/cytoimagenet}.
\end{abstract}

\section{Introduction}

Automated microscopy has given biologists a tool for studying various cell types under varying experimental conditions with minimal assistance. In recent years, automated image analysis has proven useful for drug discovery and functional genomics research; identifying the function of novel genes and cellular response to various drugs [1]. With the capacity to perform large-scale screens rapidly, tens of thousands of images are being collected daily, demanding computational methods to extract biology from these images [2]. 

Methods for automated image analysis remains an active area of research. One strategy for automating the analysis of microscopy images is to use low-level features from classic computer vision strategies; extracting thousands of general features related to area, shape, texture and intensity of single-cells [3, 4]. However, these features usually require extensive engineering, parameter tuning, and effective single-cell segmentation, all of which must be customized to the images of interest. Furthermore, these features may not capture relevant biological signal and instead capture experimental noise [5]. To reduce labor and improve the sensitivity of extracted features, some studies rely on deep learning to automatically learn features. These strategies range from supervised training with labels, autoencoders to more recently, self-supervised learning [5, 6, 7]. However, these models require users to train on their own data, which is computationally expensive, time-consuming, and requires specialized hardware like GPUs.

As an alternative to training such models, some studies instead rely on transfer learning. Most commonly, studies re-use features from neural networks trained to classify images from ImageNet, a large-scale natural image classification dataset [8, 9, 10]. By training models to perform well on large datasets, they can learn useful features that may be transferred to other datasets. 

ImageNet is a large-scale dataset of more than 14 million diverse and annotated natural images [11]. A defining factor in the success of ImageNet is their diversity of image labels and image sources. Models that perform the best on ImageNet-1K become the standard for transfer learning [12]. These models learn a diverse set of features that are able to capture complex and generalizable patterns inherent in their dataset. Surprisingly, ImageNet features have even shown good transferability on microscopy classification tasks [13, 14]. 

However, a recent study suggests that “ImageNet features are less general than previously suggested,” and that pretraining on datasets with a closer domain match to the downstream task benefits transfer learning [12]. In fact, Caicedo et al. showed that pretraining convolutional networks (CNN) on weakly labeled microscopy images yields features that outperform ImageNet features on downstream classification tasks [5]. We hypothesize that features from pretraining on a diverse large-scale microscopy dataset will yield more domain-relevant features than ImageNet. 

We present CytoImageNet, a large-scale image dataset of weakly labeled microscopy images (890K images, 894 classes). Incorporating image data and labels from 40 distinct microscopy datasets, CytoImageNet mimics the diverse and complex nature of ImageNet. We provide a framework for curating similar large-scale pretraining datasets using openly available microscopy datasets. 

We believe that pretraining on CytoImageNet may yield biologically-meaningful image representations useful in any downstream biological task. Since concatenation of CytoImageNet-trained features and ImageNet-trained features performs significantly better than either one alone, we believe that CytoImageNet pretraining provides features that are useful, and different from ImageNet features.

\section{Methods}
\label{gen_inst}

CytoImageNet is a dataset of 890,737 microscopy images sourced from 40 openly available datasets. The selected datasets originated from 5 databases: (1) Recursion, (2) Image Data Resource (IDR) [15], (3) Broad Bioimage Benchmark Collection (BBBC) [16], (4) Kaggle and (5) Cell Image Library (CIL). Electron microscopy and histopathology images were excluded. Code for the methods below is available at \url{https://github.com/stan-hua/CytoImageNet}. The dataset is available at \url{https://www.kaggle.com/stanleyhua/cytoimagenet}.




\begin{figure}
  \centering
  \fbox{\includegraphics[width=9cm, height=10.2577cm]{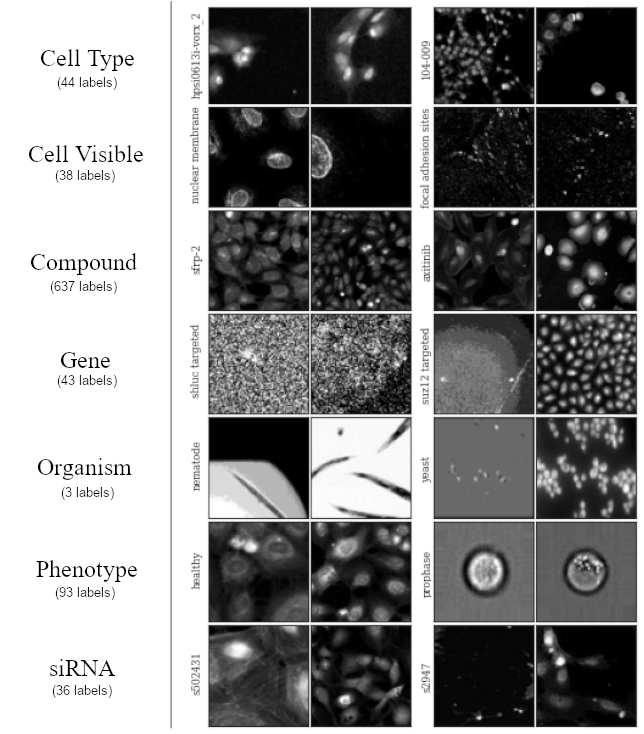}}
  \caption{Examples of CytoImageNet classes. Two representative images from two representative classes are shown for each metadata category (names are in vertical text)}
\end{figure}

\begin{table}
\centering
\caption{Number of labels contributed by each image database}
\label{tab:my-table1}
\begin{tabular}{@{}lc@{}}
\toprule
\textbf{Database}                            & \textbf{Number of Labels Contributed} \\ \midrule
Recursion                           & 651                          \\
Image Data Resource                 & 450                          \\
Broad Bioimage Benchmark Collection & 202                          \\
Kaggle                              & 27                           \\
Cell Image Library                  & 1                            \\ \bottomrule
\end{tabular}
\end{table}

\subsection{Weak label assignment \& stratified downsampling}

As a training target for CytoImageNet, we sought to associate each microscopy image with a class label. As expert-assigned annotations are not readily available or standardized across publicly available image datasets, we instead chose to assign images to classes based upon metadata associated with each of the image datasets we used. We considered 7 kinds of metadata: (organism, cell\_type, cell\_visible, phenotype, compound, gene, and sirna). Each column was searched for unique values and their counts to find potential weak labels. To create 894 labels, unique values with at least 287 images served as potential labels. Beginning with the least counts, each potential label was iterated through, searching for images that can be assigned to the label. Images that were already assigned labels are excluded, and their unique image index is stored in a hash table pointing to their label. To improve the diversity of images selected for a label, stratified sampling was done by sampling from images that were grouped by values in the remaining potential label columns.

\subsection{Image standardization \& upsampling}

To preprocess images, we converted all images to PNG. RGB images are converted to grayscale by averaging over channels. We normalized each channel between [0, 255] using the 0.1 and 99.9th percentile pixel intensity as the lower and upper ranges respectively [6]. Finally, for fluorescent images, we merged the channels. 

We took 1-4 crops from each image of varying spatial scales where possible. Unless the image was smaller than 70x70 pixels, we cropped images into four quadrants. To introduce variations in spatial scale, we further cropped these quadrants by sampling crops of either full, 0.5, 0.25, or 0.125 of the quadrant (excluding any crop scales that would produce images smaller than 70x70 pixels). Finally, we filtered any crops that were all black, all white, or where the 75th percentile pixel intensity is equal to 0. Crop pixel intensities were re-normalized as we previously did for full images.



\subsection{Training}

CytoImageNet is split into a training and validation set with 10\% used for validation. This yields roughly 900 training samples for each label. Images are fed in batches of 64 with random 0 to 360 degrees rotations. We train convolutional networks (EfficientNetB0) to classify one of the 894 labels, by minimizing the categorical cross-entropy loss of predictions to ground truth labels.  Randomly initialized models were trained for 24 epochs (2 weeks) on an NVIDIA Tesla K40C. The loss was optimized via the Adam optimizer with learning rate of 0.001.

\subsection{Evaluation}

We compare features extracted from a CytoImageNet-pretrained model, an ImageNet-pretrained model, and a randomly-initialized model on three distinct downstream tasks: 1) BBBC021 mechanism-of-action classification, 2) Cells-Out-Of-Sample (COOS-7) protein localization classification, and 3) a CYCLoPs dataset for protein localization classification [17, 18, 19]. Additionally, we would like to know if CytoImageNet and ImageNet pretraining lead to different learnt features. If so, fusing the extracted features from both may increase performance on downstream tasks. To test this, we concatenate CytoImageNet and ImageNet extracted features for each image in the transfer tasks.

BBBC021 is a dataset of fully imaged human cells. Cells are treated with one of 113 small molecules at 8 concentrations, and fluorescent images are captured staining for nucleus, actin and microtubules. The phenotypic profiling problem is presented, where the goal is to extract features containing meaningful information about the cellular phenotype exhibited. Each of 103 unique compound-concentration treatment is labeled with a mechanism-of-action (MOA). The MOA is predicted for each unique treatment (averaging features over all treatment examples) by matching the MOA of the closest point excluding points of the same compound. We use a 1-nearest neighbors using cosine distance as the metric and report the not-same-compound (NSC) accuracy.

On the other hand, the WT2 dataset from the CYCLoPs database consists of 27,058 single-cell images of yeast cells. The task is to classify the subcellular localization of a fluoresced protein, given two channels staining for the protein of interest and the cytosol. An 11-nearest neighbors is used to perform leave-one-out classification, and the accuracy is reported.

Lastly, COOS-7 contains 132,209 single-cell images of mouse cells. Images are spread over 1 training set and 4 testing sets, where each single-cell image contains a protein and nucleus fluorescent channels. COOS-7 was curated, such that each succeeding testing set has a greater degree of covariate shift from the training set. Similar to the CYCLoPs dataset, the COOS-7 dataset presents a protein subcellular localization classification task. We use a 11-nearest neighbors fit on the training set to predict on the testing sets and report the accuracy.

We chose these datasets (1) to assess the generalizability of CytoImageNet-pretrained models on distinct biological image datasets and problems, and (2) to compare against ImageNet features that have already been shown to perform relatively well on these datasets [13, 7, 19]. 

Features are extracted from the penultimate layer of EfficientNetB0 models. ImageNet-pretrained EfficietNetB0 weights are loaded in using Keras, but the weights themselves belong to the source paper [20]. Predictions are generated using k-Nearest Neighbors (kNN), and accuracy is reported with a 95\% confidence interval. BBBC021 kNN is implemented with scikit-learn, while kNN is implemented using the faiss library for COOS-7 and CYCLoPs datasets.

We compare the performance of features extracted following two optional preprocessing methods: normalization and channel merging. Normalization refers to channel normalization with the lower bound and upper bound of the 0.1 and 99.9th percentile pixel intensity respectively [6]. Channel merging refers to the merging of fluorescent channels into a single grayscale image before extracting features. If not, features are extracted for each fluorescent channel then concatenated.

\begin{table}
\centering
\caption{kNN classification accuracy (with 95\% confidence intervals) on downstream tasks, comparing EfficientNetB0 penultimate layer features. 'Random' refers to the model with randomly-initialized weights. 'Fusion' refers to the concatenation of ImageNet features and CytoImageNet features. Concatenation of randomly-initialized features with ImageNet features gave no increase in performance (data not shown).}
\label{tab:my-table2}
\begin{tabular}{ccccc}
\hline
\cellcolor[HTML]{FFFFFF}{\color[HTML]{000000} Downstream Task} & Random & ImageNet  & CytoImageNet & Fusion\\ \hline
BBC021           & 27.18 $\pm$ 8.59\% & \textbf{83.5} $\pm$ 7.17\%  & \textbf{83.5} $\pm$ 7.17\% & \textbf{86.41} $\pm$ 6.62\% \\ \hline
CYCLoPs          & 53.06 $\pm$ 0.59\% & \textbf{68.47} $\pm$ 0.55\% & 65.19 $\pm$ 0.57\% & \textbf{77.97} $\pm$ 0.49\%         \\ \hline
COOS7 Test Set 1 & 65.68 $\pm$ 0.91\% & \textbf{88.87} $\pm$ 0.61\% & 88.58 $\pm$ 0.61\% & \textbf{94.80} $\pm$ 0.43\%         \\ \hline
COOS7 Test Set 2 & 67.81 $\pm$ 0.7\%  & 88.93 $\pm$ 0.47\%          & \textbf{89.37} $\pm$ 0.46\% & \textbf{95.07} $\pm$ 0.33\% \\ \hline
COOS7 Test Set 3 & 48.77 $\pm$ 0.54\% & \textbf{75.91} $\pm$ 0.46\% & 65.97 $\pm$ 0.51\% & \textbf{78.97} $\pm$ 0.44\%         \\ \hline
COOS7 Test Set 4 & 51.88 $\pm$ 0.56\% & \textbf{82.19} $\pm$ 0.43\% & 78.52 $\pm$ 0.46\%  & \textbf{87.47} $\pm$ 0.37\%        \\ \hline
\end{tabular}
\end{table}

\section{Discussion}

CytoImageNet pretraining shows comparable performance to ImageNet pretraining on all downstream  tasks. Normalizing pixel intensities for each channel and extracting features for each channel led to the best performance for all features, implying that automated methods for feature extraction still require some form of image preprocessing. 

Given that CytoImageNet has a closer domain match to the downstream microscopy tasks, it is surprising that CytoImageNet features alone do not outperform ImageNet features. A recent study on ImageNet pretraining reported a strong correlation between ImageNet validation accuracy and transfer accuracy [12]. Our trained EfficientNetB0 model only achieved 13.42\% accuracy on the training set and more importantly, 11.32\%  on the CytoImageNet validation set, where 402 of the 894 labels had no images correctly predicted. Of which, 375 of the 402 were compound labels. Meanwhile, EfficientNetB0 was shown to attain a top-1 accuracy of 77.1\% on ImageNet-1K [20]. Given our low validation accuracy, we speculate that hyperparameter optimization will lead to further increases in benchmark classification accuracy. We believe this is an important area for future research.


Encouragingly, fusing extracted CytoImageNet and ImageNet features (via concatenation) performs the best on downstream tasks by a significant margin. This implies that CytoImageNet and ImageNet pretraining result in the learning of different albeit meaningful image representations. We believe this is due to CytoImageNet-trained models learning some feature of microscopy images that are not available in natural images. These results suggest that future applications of bioimage transfer learning may benefit the most from the fusion of CytoImageNet and ImageNet features.


In summary, we present CytoImageNet; a pretraining dataset of 890,737 microscopy images and 894 classes. We contribute a simple framework for combining openly available microscopy datasets, in the hopes that this sparks interest in the vast amount of rich microscopy data available online. Furthermore, we hope that this may inspire others to experiment on their own; both to create new large-scale datasets and to develop image classification architectures and training procedures that generalize to other image domains.

\section*{Acknowledgements}

Stanley Z. Hua was funded by the University of Toronto Cell \& Systems Biology undergraduate research award. Alan M. Moses holds a Tier II Canada Research Chair. The authors would like to acknowledge Recursion, Image Data Resource, Broad Bioimage Benchmark Collection and Kaggle for allowing open access to their microscopy datasets, which were used to construct the CytoImageNet dataset. We would like to thank the authors responsible for uploading their data on these databases. We would also like to thank Juan Caicedo and Anne Carpenter for their valuable input regarding the Broad Institute datasets. Finally, the authors would like to acknowledge the Canadian Foundation for Innovation that sponsored the NVIDIA GPU with which the models were trained on.

\section*{References}

[1] Caicedo JC, Singh S, Carpenter AE. Applications in image-based profiling of perturbations. Current Opinion in Biotechnology. 2016;39:134–142. doi:10.1016/j.copbio.2016.04.003 

[2] Wollman R, Stuurman N. High throughput microscopy: From raw images to discoveries. Journal of Cell Science. 2007;120(21):3715–3722. doi:10.1242/jcs.013623 

[3] Danuser G. Computer Vision in cell biology. Cell. 2011;147(5):973–978. doi:10.1016/j.cell.2011.11.001 

[4] Carpenter AE, Jones TR, Lamprecht MR, Clarke C, Kang I, Friman O, Guertin DA, Chang J, Lindquist RA, Moffat J, et al. CellProfiler: image analysis software for identifying and quantifying cell phenotypes. Genome Biology. 2006;7(10). doi:10.1186/gb-2006-7-10-r100 

[5] Caicedo JC, McQuin C, Goodman A, Singh S, Carpenter AE. Weakly supervised learning of single-cell feature embeddings. 2018 IEEE/CVF Conference on Computer Vision and Pattern Recognition. 2018. doi:10.1109/cvpr.2018.00970 

[6] Kraus OZ, Grys BT, Ba J, Chong Y, Frey BJ, Boone C, Andrews BJ. Automated Analysis of high‐content microscopy data with Deep Learning. Molecular Systems Biology. 2017;13(4):924. doi:10.15252/msb.20177551 

[7] Lu AX, Kraus OZ, Cooper S, Moses AM. Learning unsupervised feature representations for single cell microscopy images with paired cell inpainting. PLOS Computational Biology. 2019;15(9). doi:10.1371/journal.pcbi.1007348 

[8] Huh M, Agrawal P, Efros AA. What makes ImageNet good for transfer learning? arXiv. 2016;1608.08614. 

[9] Razavian AS, Azizpour H, Sullivan J, and Carlsson S. CNN features off-the-shelf: an astounding baseline for recognition. In IEEE Conference on Computer Vision and Pattern Recognition Workshops (CVPRW),pages 512–519. IEEE, 2014.

[10] Donahue J, Jia Y, Vinyals O, Hoffman J, Zhang N, Tzeng E, and Darrell T. Decaf: A deep convolutional activation feature for generic visual recognition. In International Conference on Machine Learning, pages 647–655, 2014.

[11] Deng J, Dong W, Socher R, Li L-J, Kai Li, Li Fei-Fei. ImageNet: A large-scale hierarchical image database. 2009 IEEE Conference on Computer Vision and Pattern Recognition. 2009. doi:10.1109/cvpr.2009.5206848 

[12] Kornblith S, Shlens J, Le QV. Do Better Imagenet Models Transfer Better? 2019 IEEE/CVF Conference on Computer Vision and Pattern Recognition (CVPR). 2019. doi:10.1109/cvpr.2019.00277 

[13] Pawlowski N, Caicedo JC, Singh S, Carpenter AE, Storkey A. Automating morphological profiling with generic deep convolutional networks. bioRxiv. 2016. doi:10.1101/085118 

[14] Kensert A, Harrison PJ, Spjuth O. Transfer learning with deep convolutional neural networks for classifying cellular morphological changes. SLAS DISCOVERY: Advancing the Science of Drug Discovery. 2019;24(4):466–475. doi:10.1177/2472555218818756 

[15] Williams E, Moore J, Li SW, Rustici G, Tarkowska A, Chessel A, Leo S, Antal B, Ferguson RK, Sarkans U, et al. Image Data Resource: A Bioimage Data Integration and publication platform. Nature Methods. 2017;14(8):775–781. doi:10.1038/nmeth.4326 

[16] Ljosa V, Sokolnicki KL, Carpenter AE. Annotated high-throughput microscopy image sets for validation. Nature Methods. 2012;9(7):637–637. doi:10.1038/nmeth.2083

[17] Caie PD, Walls RE, Ingleston-Orme A, Daya S, Houslay T, Eagle R, Roberts ME, Carragher NO. High-content phenotypic profiling of drug response signatures across distinct cancer cells. Molecular Cancer Therapeutics. 2010;9(6):1913–1926. doi:10.1158/1535-7163.mct-09-1148 

[18] Koh JL, Chong YT, Friesen H, Moses A, Boone C, Andrews BJ, Moffat J. Cyclops: A comprehensive database constructed from automated analysis of protein abundance and subcellular localization patterns in saccharomyces cerevisiae. G3 Genes|Genomes|Genetics. 2015;5(6):1223–1232. doi:10.1534/g3.115.017830 

[19] Lu AX, Lu AX, Schormann W, Andrews DW, Moses AM. The Cells Out of Sample (COOS) dataset and benchmarks for measuring out-of-sample generalization of image classifiers. ArXiv. 2019;abs/1906.07282. 

[20] Tan M, Le Q. EfficientNet: Rethinking Model Scaling for Convolutional Neural Networks. In: 36th International Conference on Machine Learning (ICML 2019) Long Beach, California, USA, 9-15 June 2019. Red Hook, NY: Curran Associates, Inc; 2019. 

[21] Zhang H, Cao X, Tang M, Zhong G, Si Y, Li H, Zhu F, Liao Q, Li L, Zhao J, et al. A subcellular map of the human kinome. eLife. 2021;10. doi:10.7554/elife.64943 

\newpage

\section* {Supplementary Material}

\setcounter{table}{0}
\setcounter{subsection}{0}
\renewcommand{\tablename}{Supplementary Table}

\begin{table}[hbt!]
\centering
\caption{List of Datasets Used with External Links}
\label{tab:my-table3}
\resizebox{\textwidth}{!}{%
\begin{tabular}{@{}llll@{}}
\toprule
\textbf{Database} & \textbf{Shorthand} & \textbf{Dataset Name}                                & \textbf{Link}                                              \\ \midrule
BBBC              & bbbc012            & C. elegans infection marker                          & https://bbbc.broadinstitute.org/BBBC012                    \\ \midrule
BBBC              & bbbc010            & C. elegans live/dead assay                           & https://bbbc.broadinstitute.org/BBBC010                    \\ \midrule
BBBC              & bbbc011            & C. elegans metabolism assay                          & https://bbbc.broadinstitute.org/BBBC011                    \\ \midrule
BBBC              & bbbc048            & Cell Cycle Jurkat Cells                              & https://bbbc.broadinstitute.org/BBBC048                    \\ \midrule
BBBC              & bbbc017            & Human HT29 colon-cancer cells shRNAi screen          & https://bbbc.broadinstitute.org/BBBC017                    \\ \midrule
BBBC   & bbbc026          & Human Hepatocyte and Murine Fibroblast cells - Co-culture experiment & https://bbbc.broadinstitute.org/BBBC026                                \\ \midrule
BBBC              & bbbc006            & Human U2OS cells (out of focus)                      & https://bbbc.broadinstitute.org/BBBC006                    \\ \midrule
BBBC              & bbbc022            & Human U2OS cells - compound cell-painting experiment & https://bbbc.broadinstitute.org/BBBC022                    \\ \midrule
BBBC              & bbbc013            & Human U2OS cells cytoplasm-nucleus translocation     & https://bbbc.broadinstitute.org/BBBC013                    \\ \midrule
BBBC              & bbbc014            & Human U2OS cells cytoplasm-nucleus translocation (2) & https://bbbc.broadinstitute.org/BBBC014                    \\ \midrule
BBBC              & bbbc015            & Human U2OS cells transfluor                          & https://bbbc.broadinstitute.org/BBBC015                    \\ \midrule
BBBC              & bbbc025            & Human U2OS cells - RNAi Cell Painting experiment     & https://bbbc.broadinstitute.org/BBBC025                    \\ \midrule
BBBC              & bbbc045            & Human White Blood Cells                              & https://bbbc.broadinstitute.org/BBBC045                    \\ \midrule
BBBC              & bbbc051            & Human kidney cortex cells                            & https://bbbc.broadinstitute.org/BBBC051                    \\ \midrule
BBBC              & bbbc038            & Kaggle 2018 Data Science Bowl                        & https://bbbc.broadinstitute.org/BBBC038                    \\ \midrule
BBBC              & bbbc020            & Murine bone-marrow derived macrophages               & https://bbbc.broadinstitute.org/BBBC020                    \\ \midrule
BBBC              & bbbc041            & P. vivax (malaria) infected human blood smears       & https://bbbc.broadinstitute.org/BBBC041                    \\ \midrule
BBBC              & bbbc005            & Synthetic cells                                      & https://bbbc.broadinstitute.org/BBBC005                    \\ \midrule
CIL               & cil\_kinome        & Kinome Atlas                                         & http://cellimagelibrary.org/pages/kinome\_atlas            \\ \midrule
IDR               & idr0081            & Adenovirus                                           & https://idr.openmicroscopy.org/webclient/?show=screen-2406 \\ \midrule
IDR               & idr0017            & Chemical-Genetic Interaction Map                     & https://idr.openmicroscopy.org/webclient/?show=screen-1151 \\ \midrule
IDR               & idr0016            & Compound Profiling                                   & https://idr.openmicroscopy.org/webclient/?show=screen-1251 \\ \midrule
IDR               & idr0009            & Early Secretory Pathway                              & https://idr.openmicroscopy.org/webclient/?show=screen-803  \\ \midrule
IDR               & idr0041            & Mitotic Atlas                                        & https://idr.openmicroscopy.org/webclient/?show=project-404 \\ \midrule
IDR               & idr0021            & Pericentriolar Material                              & https://idr.openmicroscopy.org/webclient/?show=project-51  \\ \midrule
IDR               & idr0080            & Perturbation                                         & https://idr.openmicroscopy.org/webclient/?show=screen-2701 \\ \midrule
IDR               & idr0088            & Phenomic Profiling                                   & https://idr.openmicroscopy.org/webclient/?show=screen-2651 \\ \midrule
IDR               & idr0003            & Plasticity                                           & https://idr.openmicroscopy.org/webclient/?show=screen-51   \\ \midrule
IDR               & idr0072            & Subcellular Localization                             & https://idr.openmicroscopy.org/webclient/?show=screen-2952 \\ \midrule
IDR               & idr0037            & Variation in Human iPSC lines                        & https://idr.openmicroscopy.org/webclient/?show=screen-2051 \\ \midrule
IDR               & idr0067            & Yeast Meiosis                                        & https://idr.openmicroscopy.org/webclient/?show=project-904 \\ \midrule
IDR    & idr0093          & siRNA screen for cell size and RNA production                        & https://idr.openmicroscopy.org/webclient/?show=screen-2751             \\ \midrule
Kaggle & kag\_cell\_cycle & Cell Cycle Experiments                                               & https://www.kaggle.com/paultimothymooney/cell-cycle-experiments        \\ \midrule
Kaggle & kag\_hpa\_single & Human Protein Atlas - Single Cell Classification                     & https://www.kaggle.com/c/hpa-single-cell-image-classification/data     \\ \midrule
Kaggle & kag\_hpa         & Human Protein Atlas Image Classification                             & https://www.kaggle.com/c/human-protein-atlas-image-classification/data \\ \midrule
Kaggle            & kag\_leukemia      & Leukemia Classification                              & https://www.kaggle.com/andrewmvd/leukemia-classification   \\ \midrule
Recursion         & rec\_rxrx1         & RxRx1                                                & https://www.rxrx.ai/rxrx1\#the-data                        \\ \midrule
Recursion         & rec\_rxrx19a       & RxRx19a                                              & https://www.rxrx.ai/rxrx19a                                \\ \midrule
Recursion         & rec\_rxrx19b       & RxRx19b                                              & https://www.rxrx.ai/rxrx19b                                \\ \midrule
Recursion         & rec\_rxrx2         & RxRx2                                                & https://www.rxrx.ai/rxrx2                                  \\ \bottomrule
\end{tabular}%
}
\end{table}

\end{document}